\documentclass[letterpaper, 10 pt, conference]{ieeeconf}  %

\IEEEoverridecommandlockouts                              %

\usepackage{times} %

\usepackage{comment}
\usepackage[T1]{fontenc}
\usepackage{amssymb}
\usepackage{mathtools}

\usepackage{booktabs}
\usepackage{makecell}
\usepackage[normalem]{ulem}
\useunder{\uline}{\ul}{}
\usepackage{multirow}
\usepackage[dvipsnames]{xcolor}
\definecolor{enumcolor}{HTML}{6a2ea3}
\usepackage{tikz}
\newcommand{\circlednum}[2][enumcolor]{\strut\raisebox{-1pt}{\tikz{\node[circle,inner sep=0.6pt,fill=#1,font=\scriptsize\bfseries\color{white}]{#2};}}}
\usepackage{censor}
\usepackage{hyperref}
\usepackage[absolute]{textpos}

\usepackage[capitalize]{cleveref}
\crefname{section}{Sec.}{Secs.}
\Crefname{section}{Section}{Sections}
\Crefname{table}{Table}{Tables}
\crefname{table}{Tab.}{Tabs.}

\title{\LARGE \bf
Spiking Patches:\\Asynchronous, Sparse, and Efficient Tokens for Event Cameras
}

\author{Christoffer Koo Øhrstrøm and Ronja Güldenring and Lazaros Nalpantidis%
\thanks{This work has been supported by Innovation Fund Denmark through the project ``Safety and Autonomy for Vehicles in Agriculture (SAVA)", 2105-00013A.}
\thanks{All authors are with the Department of Electrical and Photonics Engineering, DTU - Technical University of Denmark, Kgs. Lyngby, Denmark.
Corresponding author: {\tt\footnotesize chkoo@dtu.dk}}
}

\begin{document}

\definecolor{somegray}{rgb}{0.5, 0.5, 0.5}
\newcommand{\darkgrayed}[1]{\textcolor{somegray}{#1}}
\begin{textblock}{10}(3, 0.5)
\begin{center}
\darkgrayed{This paper has been accepted for publication at the \\
IEEE/RSJ International Conference on Intelligent Robots \& Systems (IROS), 2026}
\end{center}
\end{textblock}

\maketitle
\thispagestyle{empty}
\pagestyle{empty}

\begin{abstract}
Event cameras are becoming increasingly appealing for robotic applications.
Even though the output of an event camera is an asynchronous stream of spatially sparse events, prior works have represented events as synchronous dense frames or as synchronous voxels.
Given a stream of asynchronous and spatially sparse events, our goal is to discover an event representation that preserves these properties.
We propose tokenization of events and present a tokenizer, \emph{Spiking Patches}, specifically designed for event cameras.
Spiking Patches gives the means to preserve the unique properties of event cameras and we show in our experiments that this comes without sacrificing accuracy.
We evaluate our tokenizer using a Graph Neural Network, a Point Cloud Network and a Transformer on gesture recognition and object detection.
Tokens from Spiking Patches yield inference times that are up to 3.4$\times$ faster than voxel-based tokens and up to 10.4$\times$ faster than frames.
We achieve this while matching their accuracy and even surpassing in some cases with absolute improvements up to 3.8 for gesture recognition and up to 1.4 for object detection.
Code is available at \href{https://github.com/DTU-PAS/spiking-patches}{https://github.com/DTU-PAS/spiking-patches}
\end{abstract}

\section{Introduction}
\label{sec:intro}
Event cameras produce asynchronous and spatially sparse output that is inherently different from the synchronous and dense images that are typically used in robotic applications.
In this work, we propose a new event representation that is compatible with existing model architectures (Graph Neural Networks, Point Clouds Networks, and Transformers), and preserves asynchrony and spatial sparsity.
For this purpose, we propose a new tokenizer, \emph{Spiking Patches}.

The output of event cameras is an asynchronous stream of spatially sparse events.
Asynchrony and spatial sparsity are useful properties for robots and autonomous vehicles. 
For example, objects may appear suddenly and unpredictably like in autonomous driving, where potentially fast-moving objects may appear from blind spots.
Synchronous methods have a fixed reaction time, which can be too slow and result in safety issues in such situations.
However, asynchrony makes it possible to react to sudden objects as they appear \cite{gehrigLowlatencyAutomotiveVision2024}.
Likewise, spatial sparsity is useful because it allows shorter inference times by limiting computations to regions within the field of view that exhibit activity.

\begin{figure}
    \centering
    \includegraphics[width=\linewidth]{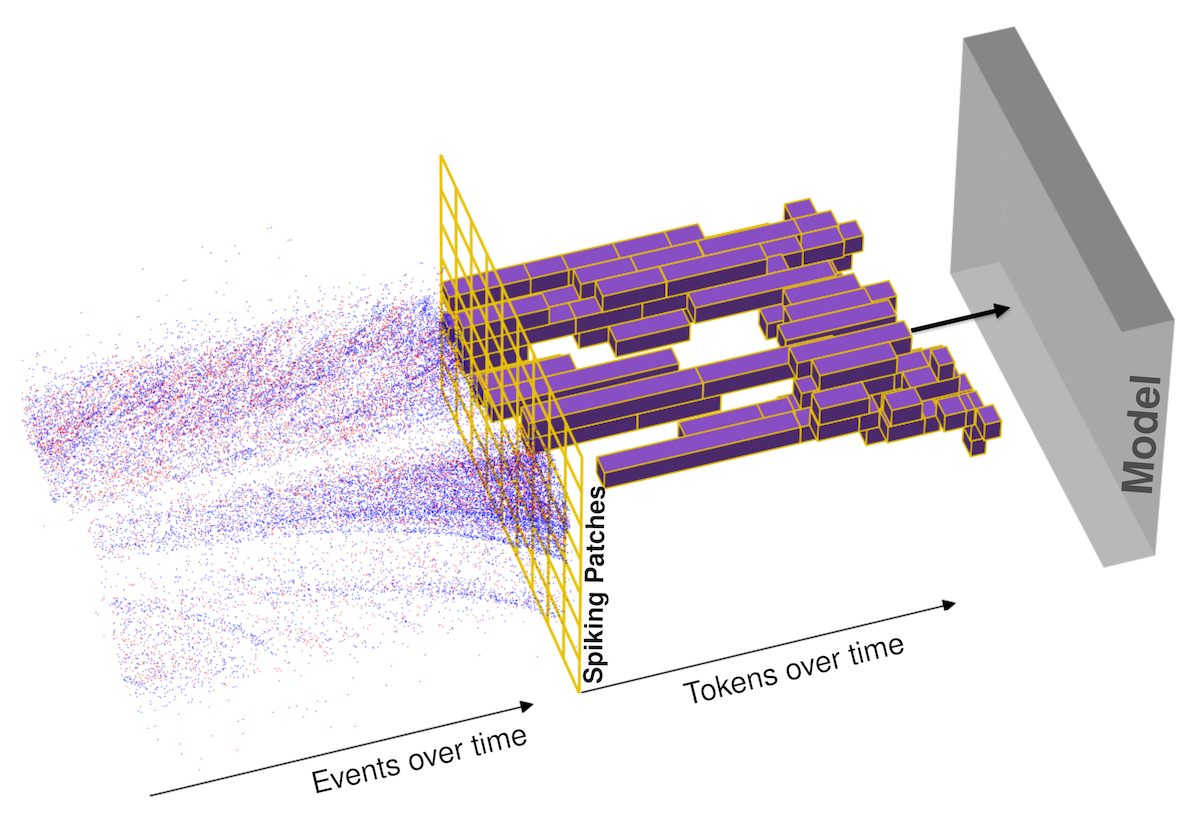}
    \caption{
    We present a tokenizer particularly developed for event cameras: \textit{Spiking Patches (SP)}.
    Events (left) are tokenized using SP (middle) and the resulting tokens (right) are processed by a model.
    We show that SP is compatible with Graph Neural Networks (GNNs), Point Cloud Networks (PCNs), and Transformers.
    SP produces tokens that preserve two unique characteristics of event cameras: Asynchrony and spatial sparsity.}
    \label{fig:intro}
\end{figure}

Despite of asynchrony and spatial sparsity being useful, many deep learning perception approaches opt for first converting the events to a frame-based representation.
This option is attractive because one can apply existing computer vision methods (e.g. CNNs and Transformers) for conventional images to events.
However, this comes at the cost of frames being constructed from a fixed time interval, resulting in a synchronous and spatially dense representation.
There are methods to mitigate this \cite{liAsynchronousSpatioTemporalMemory2022, sabaterEventTransformerSparseAware2022, sabaterEventTransformerMultiPurpose2023, pengSceneAdaptiveSparse2024, wangExploitingSpatialSparsity2022}; 
however, such methods seek to reclaim asynchrony and spatial sparsity after the frame conversion has already happened.
We argue that it is beneficial to use a representation that preserves the two properties, such that any model that uses the representation will also preserve them by default.
Currently, there is no \emph{representation} capable of preserving asynchrony and spatial sparsity, which motivates us to design such a representation.
To this end, we turn our focus to tokenization.
We present a spatio-temporal tokenizer suitable for events.
The tokens can be processed by different types of architectures such as Graph Neural Networks (GNNs), Point Cloud Networks (PCNs), and Transformers.
The tokenizer also preserves asynchrony and spatial sparsity.
We achieve this through two key observations:
(a) Events contain very little information by themselves, but groups of events contain a high amount of information, similar to how a single pixel carries little information, but a patch of pixels contains meaningful information \cite{dosovitskiyImageWorth16x162020}.
(b) Events exist in a spatio-temporal space and thus need to be grouped accordingly to both dimensions.

Our key finding is that we can tokenize events through spatial and temporal relations simultaneously by treating a patch as a spiking neuron.
We denote our tokenization method as Spiking Patches, which is illustrated in \cref{fig:intro}.

Our contributions are summarized as follows:
\\[0.25cm]
\circlednum{1} \textcolor{enumcolor}{\textbf{Spiking Patches.}} We represent events as tokens and introduce a tokenizer for event cameras.
\\[0.25cm]
\circlednum{2} \textcolor{enumcolor}{\textbf{Accuracy \& Speed.}} Spiking Patches is compared against frames and voxels for a GNN, PCN, and Transformer on gesture recognition and object detection.
We find that Spiking Patches consistently improves inference speed while matching or surpassing the accuracy of frames and voxels.
\\[0.25cm]
\circlednum{3} \textcolor{enumcolor}{\textbf{Asynchrony \& Spatial Sparsity.}} We show that Spiking Patches obtains comparable levels of asynchrony and spatial sparsity to events.
\\[0.25cm]
\circlednum{4} \textcolor{enumcolor}{\textbf{Efficiency.}} Spiking Patches reduces input sizes by at least $1.5 \times$ over frames and voxels, and we show that our implementation creates tokens faster than events arrive.

\section{Related Works}
\label{sec:related}
Our motivation for Spiking Patches is to develop a representation that preserves asynchrony and spatial sparsity.
Therefore, we relate Spiking Patches to works that have sought to preserve one or both of those properties.
\\[0.22cm]
\textbf{Frame-based methods} use conventional deep learning methods on frame representations of events.
Frame conversion breaks asynchrony and spatial sparsity, which has inspired some works to mitigate the breakage.
Event Transformer \cite{sabaterEventTransformerSparseAware2022,sabaterEventTransformerMultiPurpose2023} and DVS-ViT \cite{wangExploitingSpatialSparsity2022} increase the spatial sparsity by dividing a frame into patches and keeping active patches where the percentage of activated pixels are above a predetermined threshold.
SAST \cite{pengSceneAdaptiveSparse2024} uses learnable scoring and selection modules throughout each layer to dynamically select a sparse subset of patches.
These works share the patch-based approach of Spiking Patches, but break asynchrony because they require a fixed time interval.
The fixed time interval further has ramifications in how the methods may prematurely disregard patches in cases where slowly moving objects generate too few events to activate a patch within the time window of a frame.
We contrast this to how Spiking Patches provide controls to allow for spikes in areas of low activity.
Messikommer et al. \cite{messikommerEventBasedAsynchronousSparse2020} use a Sparse CNN \cite{graham3DSemanticSegmentation2018} to take advantage of spatial sparsity and also formulate update rules for the frame representation and Sparse CNN to make it asynchronous. However, they obtain low accuracy for object detection.
ASTMNet \cite{liAsynchronousSpatioTemporalMemory2022} adaptively samples a varying number of events to preserve asynchrony.
Spiking Patches preserves asynchrony by treating patches as spiking neurons.\\\\
\textbf{Voxels} are a generalization of 2D-patches in 3D space.
That is, they are non-overlapping and have a fixed height, width, and duration.
Voxelization can be seen as a tokenizer for event cameras \cite{liuVoxelBasedMultiScaleTransformer2024}, but voxels break asynchrony and share the same ramifications as frame-based methods wrt. the fixed time duration.
\\[0.22cm]
\textbf{Graph Neural Networks (GNNs)} have been applied to events by subsampling events as nodes and constructing edges based on the spatio-temporal distance between nodes \cite{biGraphBasedSpatioTemporalFeature2020}.
GNNs are able to fully utilize the asynchrony \cite{liGraphBasedAsynchronousEvent2021,schaeferAEGNNAsynchronousEventBased2022} and spatial sparsity of events.
Deng et al. \cite{dengVoxelGraphCNN2022} treat voxels as nodes in a graph and apply a GNN to this representation.
Spiking Patches is able to do this as well, but without breaking asynchrony as voxels do.
\\[0.22cm]
\textbf{Point Cloud Networks (PCNs).} Events are spatio-temporal streams and can be considered as 3D point clouds.
This view makes it possible to apply PCNs to events \cite{chenECSNetSpatioTemporalFeature2023,renTTPOINTTensorizedPoint2023,sekikawaEventNetAsynchronousRecursive2019}.
These methods operate directly on events and so completely preserve the unique properties of event cameras.
PCNs have shown promise in classification tasks, but are unexplored for more challenging tasks like object detection.
The tokens from Spiking Patches are a spatio-temporal stream, making them compatible with PCNs.
\\[0.22cm]
\textbf{Spiking Neural Networks (SNNs) \cite{maassNetworksSpikingNeurons1997}} and Spiking Patches share a common source of inspiration:
The biological spiking neuron.
SNNs use spiking neurons as the neural model of a neural network.
This enables SNNs to utilize the sparsity of events through spike functions in the network.
The internal model sparsity reduces the number of operations at inference.
However, this increases training difficulties because the spike function is non-differentiable and SNNs therefore require gradient approximation methods \cite{leeTrainingDeepSpiking2016,shresthaSLAYERSpikeLayer2018,wuSpatioTemporalBackpropagationTraining2018} to be trained directly.
Although SNNs promise to efficiently process sparse data such as events, they also require specialized neuromorphic hardware to fulfill this promise.
Spiking Patches avoid these complexities by applying the spike function to tokenization rather than as a part of a learnable model.

\section{Method}
\label{sec:method}

\subsection{Preliminaries}
\noindent \textbf{Events.}
Event cameras emit an event when the change in log-brightness of a pixel exceeds a threshold.
Let $L(x, y, t)$ be the log-brightness for the pixel at $(x, y)$ at time $t$, $C$ the firing threshold, and $\Delta t$ the time since the last emission for the given pixel.
The firing condition is
\begin{equation}
    \label{eqn:event_firing_condition}
    | L(x, y, t) - L(x, y, t - \Delta t) | \geq C
\end{equation}
The pixel emits an event $e = (x, y, t, p)$ when the condition in \cref{eqn:event_firing_condition} is satisfied, where $x$, $y$, and $t$ are given as in \cref{eqn:event_firing_condition} and $p \in \{ -1, 1 \}$ is the polarity of the event.
The polarity is $-1$ when the logarithmic change is negative and $1$ otherwise.
Events are an asynchronous sequence of tuples in ascending order wrt. time, which we denote as $\mathcal{E}$:
\begin{equation}
    \label{eqn:event_stream}
    \mathcal{E} = \{ e_i \}_{i=1}^{N}= \{ (x_i, y_i, t_i, p_i) \}_{i=1}^{N}
\end{equation}
\textbf{Tokenization.}
We define a tokenizer as a function from a sequence of atoms to a sequence of groups, where each group is a subsequence of atoms from the input sequence.
For example, for natural language, we could have a sequence of symbols (letters, whitespace, and punctuation) where each symbol is an atom:
\textit{[T,h,i,s, ,i,s, ,a, ,t,e,s,t,.]}.
A simple tokenizer could be to group all atoms between whitespace and punctuation:
\textit{[T,h,i,s, ,i,s, ,a, ,t,e,s,t,.] \(\rightarrow\) [This,is,a,test,.]}.
The example shows how a sequence of 15 atoms is tokenized into 5 tokens.
We notice that each atom does not carry meaning by itself, but tokens form meaningful words.
This makes it easier for models to reason about the sequence and reduces the time complexity, as the number of tokens is much smaller than the number of atoms.
These properties of tokenizers have made them ubiquitous in NLP, which motivates us to explore them for event cameras.

\subsection{Spiking Patches}
\label{sec:spiking_patches}
\begin{figure}
    \centering
    \includegraphics[width=1\linewidth]{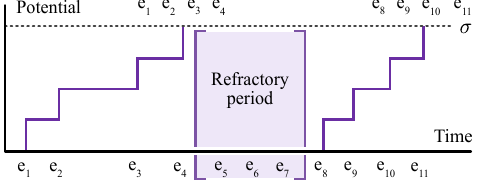}
    \caption{Overview of Spiking Patches.
    Events are arriving within a single patch.
    Each event causes a constant increase in the patch potential.
    The first spike occurs at event $e_4$ when the potential exceeds the threshold $\sigma$.
    The resulting token consists of $e_1$ through $e_4$.
    The events $e_5$ through $e_7$ are ignored as they occur within the refractory period $T$. The second spike occurs at $e_{11}$ and the resulting token consists of $e_{8}$ through $e_{11}$.}
    \label{fig:spiking_patches}
    \vspace{-0.4cm}
\end{figure}

We design our tokenizer such that each token will be spatially and temporally informative.
We can achieve the spatial relations by taking inspiration from ViT \cite{dosovitskiyImageWorth16x162020}.
That is, we divide the viewport of the event camera into a grid of non-overlapping patches (cf. \cref{fig:intro}) where each patch has height and width equal to $P$.
This division is made to restrict tokens to only include events from within the same patch.
This allows us to ensure that events within a token are spatially close to each other and therefore likely to be spatially informative.
The temporal relations in Spiking Patches are inspired by the biological spiking neuron.
We treat each patch as a spiking neuron and apply concepts such as spiking and refractory periods, as illustrated in \cref{fig:spiking_patches} and described subsequently.
\\[0.22cm]
\textbf{Threshold-based Spiking.}
Spiking neurons use spikes to send signals.
The way spiking works is that each spiking neuron has a potential, $u_i$, where $i$ is the time step.
The potential is increased when an input signal \(I_i\) arrives.
This causes a spike when $u_i \ge \sigma$ where $\sigma$ is a threshold.
The potential is reset to the resting potential $u_{rest}$ after emitting a spike.
After the reset, the potential is again increased when an input signal arrives.
The spiking mechanism is formalized in \cref{eqn:spiking_neuron_input,eqn:spiking_neuron_heaviside,eqn:spiking_neuron_update} where $H$ is the Heaviside function and $s_{i}$ denotes whether a spike occurred.
\begin{align}
    \label{eqn:spiking_neuron_input}
    v_{i} &= u_{i} + I_{i} \\
    \label{eqn:spiking_neuron_heaviside}
    s_{i} &= H(v_{i} - \sigma) \\
    \label{eqn:spiking_neuron_update}
    u_{i+1} &= s_{i} \cdot u_{rest} + (1 - s_{i}) \cdot v_{i}
\end{align}
We can adopt this spike mechanism for Spiking Patches by treating each patch as a spiking neuron.
We therefore assign a potential to each patch that we will denote as $u_{i}^{(x_p,y_p)}$.
Here, $(x_p, y_p)$ refers to the x and y locations of the patch on the grid.
We will subsequently omit the patch location for clarity.
The patch potential is updated whenever an event arrives within the area of a patch.
We use \cref{eqn:spiking_neuron_input,eqn:spiking_neuron_heaviside,eqn:spiking_neuron_update} to update the patch potential and set $I_{i} = 1$ and $u_{rest} = 0$.
Furthermore, we keep track of which events have arrived since the last spike and denote this sequence as $E_{i}$.
Once a patch exceeds the threshold \(\sigma\) at time step $i$, a new token is constructed from the events in $E_{i}$ and is assigned the spatio-temporal position $(x_p, y_p, t_{i})$ where $t_{i}$ is the time of the last event.
Finally, we reset $E_{i}$ to be the empty sequence after the token has been generated.

The spiking mechanism described above preserves asynchrony because each patch spikes independently of other patches and the generated tokens are tied to events, which themselves are asynchronous.
It also preserves spatial sparsity because tokens are only generated in patches with sufficient activity to cause spikes, which makes the resulting tokens sparse in token space.
We further validate in \cref{sec:asynchrony} and \cref{sec:spatial_sparsity} that Spiking Patches obtains asynchrony and spatial sparsity levels comparable to events.
Finally, we note that Spiking Patches is fast (validated in \cref{sec:efficiency}) with a time complexity of $O(n)$, $n$ being the number of events.
\\[0.22cm]
\textbf{Refractory Period.}
Some applications of event cameras may require control of the temporal resolution.
For example, Transformers \cite{vaswaniAttentionAllYou2017} scale quadratically with sequence length, such that fewer tokens increases model efficiency.
Again, inspired by spiking neurons, we introduce refractory periods to control the temporal resolution.
Refractory periods follow directly after a spike and have duration $T$.
Events within a refractory period have no effect on the patch potential and are not added to $E_{i}$.
In other words, events within the refractory period are discarded.
We note that the refractory period is optional and can be omitted by setting $T = 0$.
We find in \cref{sec:ablation} that a refractory period of $T = 100$ ms can reduce the input size by $4 \times$ while maintaining 96.7\% of the accuracy obtained without a refractory period.

\section{Experiments}
\begin{table*}
\centering
\caption{
Gesture Recognition on DvsGesture.
}
\begin{tabular}{@{}l|cc|cc|cc@{}}
\toprule
\multicolumn{1}{c}{} & \multicolumn{2}{c}{GNN} & \multicolumn{2}{c}{PCN} & \multicolumn{2}{c}{Transformer} \\
\midrule
Representation & Accuracy & Speed (ms) & Accuracy & Speed (ms) & Accuracy & Speed (ms) \\
\midrule
Frame & - & - & - & - & 97.0 & 28.0 \\
Voxels & \textbf{95.5} & 7.8 & \textbf{96.6} & 31.3 & 97.7 & 3.4 \\
\textbf{Spiking Patches (ours)} & \textbf{95.5} & \textbf{6.3} & \textbf{96.6} & \textbf{9.2} & \textbf{98.1} & \textbf{2.7} \\ \bottomrule
\end{tabular}
\label{tab:dvs_gesture}
\end{table*}
\begin{table*}
\caption{
Gesture Recognition on SL-Animals-DVS.
}
\centering
\begin{tabular}{@{}l|cc|cc|cc@{}}
\toprule
\multicolumn{1}{c}{} & \multicolumn{2}{c}{GNN} & \multicolumn{2}{c}{PCN} & \multicolumn{2}{c}{Transformer} \\
\midrule
Representation & Accuracy & Speed (ms) & Accuracy & Speed (ms) & Accuracy & Speed (ms) \\
\midrule
Frame & - & - & - & - & 88.0 & 26.0 \\
Voxels & 87.6 & 5.2 & \textbf{94.4} & 22.9 & 90.2 & 3.0 \\
\textbf{Spiking Patches (ours)} & \textbf{91.4} & \textbf{4.9} & \textbf{94.4} & \textbf{6.8} & \textbf{91.7} & \textbf{2.5} \\ \bottomrule
\end{tabular}
\label{tab:sl_animals_dvs}
\end{table*}
\label{sec:experiments}
The goals of the experiments are to show that \circlednum{1} Spiking  (SP) is compatible with GNNs, PCNs, and Transformers (\cref{sec:accuracy_speed}).
\circlednum{2} SP matches or surpasses the accuracy of frames and voxels while being faster than both (\cref{sec:accuracy_speed}).
\circlednum{3} SP preserves asynchrony (\cref{sec:asynchrony}) and spatial sparsity (\cref{sec:spatial_sparsity}).
\circlednum{4} SP is efficient (\cref{sec:efficiency}).
We evaluate \circlednum{1} and \circlednum{2} on 3 datasets for gesture recognition and object detection to show that SP works for more than a single dataset and task.
We leave additional tasks for future work.

\subsection{Experimental Setup}
\label{sec:experimental_setup}
\noindent \textbf{Baselines.}
We compare to frames and voxels.
Frames are a common choice for event representation and are synchronous and spatially dense.
Voxels fall into our definition of a tokenizer and are synchronous and spatially sparse.
Furthermore, voxels are conceptually similar to Spiking Patches:
They share the same patch-based approach to spatial relations, but the temporal relations are modeled differently.
Spiking Patches uses spiking neurons to model temporal relations, whereas voxels have a constant duration.
The difference in temporal relations means that voxels break asynchrony.
It also makes them less spatially sparse (\cref{sec:spatial_sparsity}), because a voxel is generated even if it only has a single event.
The latter can be addressed by requiring that a voxel contains a minimum number of events.
We apply voxel thresholding in our analyses, but not in our experiments as we found in preliminary experiments that filtering decreases voxel accuracy.
We set voxel durations to 50 ms, as is common practice for synchronous methods \cite{gehrigRecurrentVisionTransformers2023,perotLearningDetectObjects2020}.
The frame and voxel patch sizes are the same as for Spiking Patches in order to be comparable.
\\[0.22cm]
\textbf{Models.}
We choose to use well-known representative models with publicly available code as the GNN, PCN, and Transformer models.
The GNN is AEGNN \cite{schaeferAEGNNAsynchronousEventBased2022} where each token is a node in a graph.
For the PCN, we opt for PointNet++ \cite{qiPointNetDeepHierarchical2017} with a token being a point.
The Transformer is a ViT \cite{dosovitskiyImageWorth16x162020} initialized from MAE \cite{heMaskedAutoencodersAre2022}.
The Transformer is the only model to compare against frames as GNNs and PCNs are usually not applied to frames.
For all models, we first embed each token into feature space using a Stacked Histogram \cite{gehrigEndtoEndLearningRepresentations2019} with 10 logarithmically spaced buckets for the time dimension.
We further have to make slight changes to AEGNN and ViT:
AEGNN pools nodes through voxels, making all representations use voxelization in the network.
This conflicts with the goal of comparing voxels as an input representation only.
As such, we modify it to pool nodes using farthest point sampling.
For ViT, the position encoding follows the sinusoidal encoding of \cite{vaswaniAttentionAllYou2017}, but we concatenate separate encodings for the three positional dimensions: $x$, $y$, and $t$.
For object detection, we use the YOLOX \cite{geYOLOXExceedingYOLO2021} framework with the models being feature extractors.
Inference times are measured on a RTX 4090.
\\[0.22cm]
\textbf{Training.}
We use PyTorch \cite{Ansel_PyTorch_2_Faster_2024} in 16-bit mixed precision.
The GNN and PCN have a learning rate of $10^{-3}$ whereas it is $10^{-4}$ for the Transformer.
All models are trained using Adam \cite{kingmaAdamMethodStochastic2017} as optimizer and with a OneCycle \cite{smithSuperconvergenceVeryFast2019} learning rate schedule.
We train for 30K steps with a batch size of 72 for gesture recognition and for 400K steps with a batch size of 16 for object detection.
We apply EventDrop \cite{guEventDropDataAugmentation2021} and NDA \cite{liNeuromorphicDataAugmentation2022} as augmentations for all datasets.
Refractory periods are omitted ($T = 0$) unless specified otherwise.
\\[0.22cm]
\textbf{Datasets.}
We use DvsGesture \cite{amirLowPowerFully2017} and SL-Animals-DVS \cite{vasudevanSLAnimalsDVSEventdrivenSign2022} for gesture recognition.
DvsGesture records users performing 11 different hand and arm gestures (e.g. right-hand waving) in varying lighting conditions.
Each gesture lasts about 6 seconds and the dataset comprises a total of 1,342 gestures.
SL-Animals-DVS has 59 users performing 19 different sign language signs of animals.
We use GEN1 \cite{detournemireLargeScaleEventbased2020} for object detection.
GEN1 has more than 39 hours of egocentric driving using a $304 \times 240$ event camera.
It has more than 255K bounding boxes of cars and pedestrians.
\\[0.22cm]
\textbf{Evaluation Protocol.}
The sequences of DvsGesture and SL-Animals-DVS are split into subsequences of 1s and we get the final sequence-level prediction by averaging the confidences of each subsequence.
Frames use a subsequence duration of 100 ms as frames are ill-defined for long durations and we therefore consider it an unfair comparison to use a 1s window for frames.
SL-Animals-DVS does not have an official train-test split and we therefore follow the split proposed by \cite{baldwinTimeOrderedRecentEvent2023}.
For GEN1, we follow the common evaluation protocol of \cite{perotLearningDetectObjects2020}:
We remove bounding boxes with a side length or diagonal smaller than 10 or 30 pixels, respectively.
We use COCO mAP \cite{linMicrosoftCOCOCommon2014} as our main metric.
Predictions are made every 50 ms, such that frames and voxels have a duration of 50 ms.
Models based on Spiking Patches use the tokens that spike within the 50 ms window, but events within a token are allowed to be accumulated from across the window boundary.
This reflects how one would stream events and tokens in a real-world scenario.

\subsection{Accuracy \& Speed}
\label{sec:accuracy_speed}
\begin{table*}
\centering
\caption{
Object detection on GEN1.
}
\begin{tabular}{@{}l|cc|cc|cc@{}}
\toprule
\multicolumn{1}{c}{} & \multicolumn{2}{c}{GNN} & \multicolumn{2}{c}{PCN} & \multicolumn{2}{c}{Transformer} \\
\midrule
Representation & mAP & Speed (ms) & mAP & Speed (ms) & mAP & Speed (ms) \\
\midrule
Frame & - & - & - & - & \textbf{39.3} & 5.6 \\
Voxels & 37.7 & 5.8 & 38.0 & 7.5 & 37.9 & 5.4 \\
\textbf{Spiking Patches (ours)} & \textbf{39.1} & \textbf{5.3} & \textbf{38.4} & \textbf{6.2} & 39.2 & \textbf{5.1} \\ \bottomrule
\end{tabular}
\label{tab:gen1}
\end{table*}
\begin{table*}
\centering
\caption{
Comparison between events and Spiking Patches on gesture recognition.
}
\resizebox{\linewidth}{!}{%
\begin{tabular}{@{}l|cc|cc|cc|cc@{}}
\multicolumn{1}{c}{} & \multicolumn{4}{c}{\textbf{DvsGesture}} & \multicolumn{4}{c}{\textbf{SL-Animals-DVS}} \\
\toprule
\multicolumn{1}{c}{} & \multicolumn{2}{c}{GNN} & \multicolumn{2}{c}{PCN} & \multicolumn{2}{c}{GNN} & \multicolumn{2}{c}{PCN} \\
\midrule
Representation & Accuracy & Speed (ms) & Accuracy & Speed (ms) & Accuracy & Speed (ms) & Accuracy & Speed (ms) \\ \midrule
Events & 95.1 & 35.1 & 96.2 & 80.0 & \textbf{91.4} & 31.6 & 92.5 & 72.6 \\
\textbf{Spiking Patches (ours)} & \textbf{95.5} & \textbf{6.3} & \textbf{96.6} & \textbf{9.2} & \textbf{91.4} & \textbf{4.9} & \textbf{96.6} & \textbf{6.8} \\ \bottomrule
\end{tabular}%
}
\label{tab:events}
\end{table*}
\noindent \textbf{Gesture Recognition.}
Gesture Recognition provides an interesting setting for evaluating Spiking Patches because gestures are performed rapidly and have high spatial sparsity.
We set $P = 8$ and $\sigma = 25$ for the GNN, $P = 4$ and $\sigma = 20$ for the PCN, and $P = 16$ and $\sigma = 250$ for the Transformer.
These numbers are obtained on held-out validation sets.
\cref{tab:dvs_gesture} shows that Spiking Patches matches the accuracy of voxels for the GNN and PCN, and surpasses both frames and voxels for the Transformer by 1.1 and 0.4, respectively.
Spiking Patches also results in the fastest average inference times in all cases.
It is particularly noticeable for the PCN where Spiking Patches is 3.4x faster than a PCN with voxels, and for a Transformer where the speedup is 10.4x over frames.
We see the results on SL-Animals-DVS in \cref{tab:sl_animals_dvs}.
Spiking Patches remains the representation that yields the lowest inference times where the speedup remains 3.4x for a PCN and 10.4x for a Transformer.
The accuracy gains are bigger on SL-Animals-DVS with a significant 3.8 increase over voxels for a GNN, and respectively 3.7 and 1.5 increases over frames and voxels for a Transformer.
The gesture recognition experiments confirm Spiking Patches to be a fast and accurate representation for gesture recognition and that this holds for a GNN, PCN, and Transformer.
\\[0.22cm]
\textbf{Object Detection.}
Event-based object detection is a challenging task where frame-based methods tend to obtain the highest accuracy \cite{gehrigRecurrentVisionTransformers2023, zubicChaosComesOrder2023, pengGETGroupEvent2023}.
The GNN and PCN use $\sigma = 50$ and $P = 8$ while the Transformer uses $\sigma = 250$ and $P = 16$.
The values are obtained from optimizing on the GEN1 validation set.
We also set \(T = 25 \; \text{ms}\) for the Transformer as we find in \cref{sec:ablation_refractory} that this greatly decreases the number of tokens with only a minor decrease in mAP.
\cref{tab:gen1} shows the object detection results.
We see that Spiking Patches consistently results in the lowest inference times.
The speed improvement comes without sacrificing accuracy, as we find that Spiking Patches is merely 0.1 mAP below the accuracy of a frame representation.
We also see that Spiking Patches are more accurate than voxels, with a mAP increase of 1.4 for the GNN, 0.4 for the PCN, and 1.3 for the Transformer.
This highlights the potential of tokenization for event-based vision.
The experiments suggest the potential may hold across tasks as the high accuracy is consistent for both gesture recognition and object detection.
\\[0.22cm]
\textbf{Event Baseline.}
Our experiments focus on comparing Spiking Patches to voxels and frames.
However, it is also relevant to consider how it compares to events.
That is, we want to know whether a tokenizer improves results over not having any representation.
We evaluate this on gesture recognition for the GNN and PCN.
We omit the Transformer from this experiment, since the number of events and the quadratic complexity of Transformers make it prohibitive to apply it directly to events.
We follow previous works \cite{schaeferAEGNNAsynchronousEventBased2022} and subsample a maximum of 10K events as the GNN and PCN also become prohibitive to train on an academic budget without this restriction.
We see in \cref{tab:events} that Spiking Patches are more accurate than using events directly, with the only exception of a GNN on SL-Animals-DVS where events and Spiking Patches are equally accurate.
The accuracy increase is greatest at 4.1 for a PCN on SL-Animals-DVS.
Inference times also decrease significantly with consistent speedups from 5.6$\times$ to 10.7$\times$.
These results show the advantage of tokenization for event cameras---even on methods like GNNs and PCNs that are typically applied directly to events.

\subsection{Asynchrony}
\label{sec:asynchrony}
We are interested in preserving asynchrony because it enables low reaction times \cite{amirLowPowerFully2017}.
As such, we analyze asynchrony by how fast Spiking Patches captures the same amount of information as events do.
Information is here measured as the cumulative number of events contained in the tokens.
We choose a short 150 ms sequence from GEN1 where a motorcycle enters from the right.
\begin{figure}
    \centering
    \includegraphics[width=\linewidth]{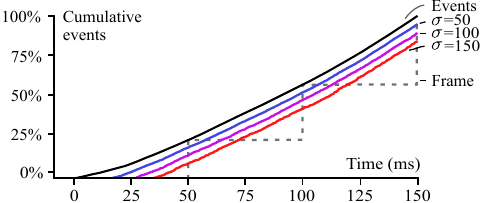}
    \caption{Event accumulation for events, Spiking Patches (3 different thresholds), and frames.
    Frames are synchronous with a 50 ms response time.
    Spiking Patches is asynchronous and follows the slope of the event curve, albeit with a 5 ms to 20 ms delay.
    Spiking Patches can react faster than synchronous methods.}
    \label{fig:asynchrony}
\end{figure}
\Cref{fig:asynchrony} shows event accumulation for events, Spiking Patches for $P = 16$ and $\sigma \in \{50, 100, 150\}$, and a synchronous frame with a 50 ms duration.
We see that Spiking Patches follows the slope of the event curve.
This shows how Spiking Patches are asynchronous and start to obtain information to react before synchronous representations like frames.
We also see that the delay between the event curve and Spiking Patches becomes larger as $\sigma$ increases.
The delay is small at around 5 ms for $\sigma = 50$.
This is a sufficiently small delay for reaction times to remain fast.
However, we also see how the delay becomes larger at around 20 ms for $\sigma = 150$.
This will still enable faster reaction times than synchronous methods, but we assess that 20 ms can be too long in safety critical scenarios.
To summarize, Spiking Patches preserves asynchrony, but $\sigma$ should be carefully chosen as to not incur too long delays.

\subsection{Spatial Sparsity}
\label{sec:spatial_sparsity}
We will here analyze how well Spiking Patches (SP) preserves spatial sparsity as compared to frames (F) and voxels (V).
We define spatial sparsity as the percentage of cells with 0 tokens in a temporal window of 50 ms.
Let this be $\text{sparsity}(\mathcal{T})$ where $\mathcal{T}$ is a tokenizer, and we set the patch size to $P = 16$ in this analysis.
The same definition follows for events:
The percentage of pixels with 0 events.
We denote this as $\text{sparsity}(\mathcal{E})$.
\begin{figure}
    \centering
    \includegraphics[width=\linewidth]{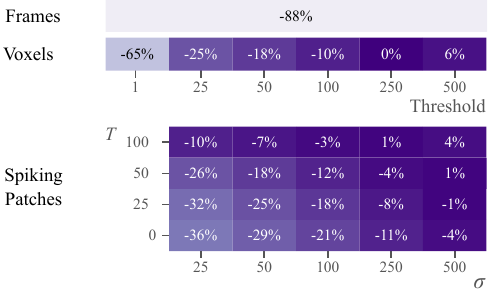}
    \caption{
    Spatial sparsity differences between events and frames (top), voxels (middle), and Spiking Patches (bottom).
    We use the colormap: \includegraphics{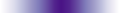}.
    Negative values are more dense and positive values are more sparse than events.
    Spiking Patches has comparable sparsity to events for $\sigma \in [250, 500]$ or $T = 100$ ms.
    }
    \label{fig:sparsity}
\end{figure}
\cref{fig:sparsity} shows the average of $\text{sparsity}(\mathcal{T}) - \text{sparsity}(\mathcal{E})$ on GEN1 for $\mathcal{T} \in \{ \text{F}, \text{V}, \text{SP} \}$ for different thresholds and refractory periods.
Negative values indicate increased density and positive values indicate increased sparsity over events.
Frames are dense and obtain the greatest difference at \(-88\%\) as \(\text{sparsity}(\mathcal{E}) = 88\%\) and \(\text{sparsity}(\text{F}) = 0\%\).
Voxels preserve spatial sparsity well for thresholds of 100 or higher with sparsity differences ranging from $-10\%$ to $6\%$.
We found in preliminary experiments that voxels are more accurate for lower thresholds.
Therefore, voxels require a low threshold of 1 to be the most accurate, but this greatly decreases the spatial sparsity with a difference of $-65\%$.
Spiking Patches preserves spatial sparsity well for $\sigma \ge 250$ or $T \ge 100 \; \text{ms}$ with sparsity differences ranging from $-11\%$ to $4\%$.
Incidentally, we find in the ablation studies (\cref{sec:ablation}) that $\sigma = 250$ is optimal for a Transformer with $P = 16$ on GEN1.
In summary, we find that Spiking Patches is able to preserve spatial sparsity.

\subsection{Efficiency}
\label{sec:efficiency}
\noindent \textbf{Input Size.}
\begin{table}
\centering
\caption{
Average input size on GEN1 50 ms windows for events, frames, voxels, and Spiking Patches for \(T \in \{0, 25, 50, 100\}\) ms.
}
\resizebox{\linewidth}{!}{
\begin{tabular}{@{}ccc|cccc@{}}
\toprule
\multirow{2}{*}{Events} & \multirow{2}{*}{Frame} & \multirow{2}{*}{Voxels} & \multicolumn{4}{c}{Spiking Patches} \\
& & & $0$ ms & $25$ ms & $50$ ms & $100$ ms \\ \midrule
35,835 & 285 & 220 & 143 & 64 & 46 & 32 \\ \bottomrule
\end{tabular}
}
\label{tab:input_sizes}
\end{table}
Closely related to spatial sparsity, we are also interested in comparing how much Spiking Patches reduces the input size.
This is of interest because a lower input size leads to faster training and inference times.
We see input size reductions in \cref{tab:input_sizes} where the number of tokens from Spiking Patches is compared to the number of inputs from events, frames (number of patches), and voxels without thresholding.
This analysis is averaged over 50 ms windows on GEN1.
Furthermore, we set \(\sigma = 250\), \(T \in \{0, 25, 50, 100\}\) ms, and set \(P = 16\).
We find that Spiking Patches reduces the input size significantly in all cases.
The smallest average reduction is 1.5$\times$ for voxels and \(T = 0\) ms.
We see the greatest reduction for events at reductions ranging from $251 \times$ to $1120 \times$ fewer tokens than events.
This is a large reduction and explains the significant speed differences for the event baseline in \cref{tab:events}.
Likewise, we find large reductions for frames where the number of tokens from Spiking Patches is on average at least $2 \times$ smaller than the number of patches in frames.
We find that Spiking Patches greatly reduces the input size across all considered representations.
\\[0.22cm]
\textbf{Tokenization Speed.}
Event cameras are particularly interesting for real-time applications.
Spiking Patches will therefore have to be fast enough to deal with such time-critical applications.
Spiking Patches are implemented in Rust to realize this goal.
Likewise, our voxel implementation uses Rust for a fair comparison.
We evaluate real-time applicability by comparing the average tokenization speed (measured as the number of events tokenized per second) and the average number of incoming events per second in GEN1.
The tokenization speed is measured using an Intel I9 CPU.
The results are:
\begin{center}
    \begin{tabular}{@{}ccc@{}}
        Events & Voxels & Spiking Patches \\ \midrule
        0.7M & 23.0M & \textbf{39.8M} \\
    \end{tabular}
\end{center}
The tokenization speed greatly exceeds the event rate for both voxels and Spiking Patches, and Spiking Patches is almost twice as fast as voxels.
We find that Spiking Patches are fast and highly relevant to real-time applications.

\subsection{Ablation Studies}
\label{sec:ablation}
\begin{figure}
    \centering
    \includegraphics[width=\linewidth]{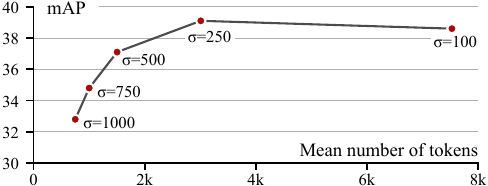}
    \caption{
    Ablation of spike threshold \(\sigma\).
    mAP remains high for $\sigma \leq 250$ and decreases afterwards.
    }
    \label{fig:spike_threshold}
\end{figure}

\begin{figure}
    \centering
    \includegraphics[width=\linewidth]{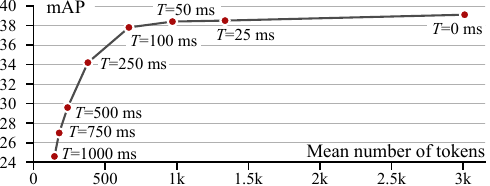}
    \caption{
    Ablation of the refractory period \(T\).
    mAP decreases slightly while the input size is reduced $4 \times$ as $T$ goes from 0 ms to 100 ms.
    \(T\) is a useful parameter to improve inference speed.
    }
    \label{fig:absolute_refractory}
\end{figure}

We are interested in studying the effect of the spike parameters and understanding the trade-off between accuracy and the mean number of tokens.
The number of tokens is interesting because it directly relates to inference time.
We measure the number of tokens as a mean count over 1s temporal windows.
The experiments are evaluated on the GEN1 \cite{detournemireLargeScaleEventbased2020} validation set and are trained for 200K steps due to computational restrictions.
All experiments use $P = 16$.
\\[0.22cm]
\textbf{Spike Threshold.}
\label{sec:ablation_spike_threshold}
\cref{fig:spike_threshold} shows how the sequence length and mAP change for different values of $\sigma$.
We observe that the largest mAP is obtained for $\sigma = 250$ and is closely followed for $\sigma = 100$.
Furthermore, mAP decreases as $\sigma$ increases beyond 250.
We also observe that the mean number of tokens for $\sigma = 250$ is significantly lower at around 3k than it is for $\sigma = 100$ at around 7.5k.
Interestingly, we see that 250 is close to the area of the patch for $P = 16$.
We hypothesize that $\sigma = P^2$ generally provides a reasonable trade-off.
\\[0.22cm]
\textbf{Refractory Periods.}
\label{sec:ablation_refractory}
\cref{fig:absolute_refractory} shows how mAP and the mean number of tokens change for $\sigma = 250$ as the duration of the refractory period increases.
We observe slightly small drops in mAP for $T \in [0, 100] \; \text{ms}$, but with increasingly large reductions in the number of tokens.
The number of tokens is reduced by \(4 \times\) from $T = 0 \; \text{ms}$ to \(T = 100 \; \text{ms}\), while mAP decreases from 39.1 to 37.8, and thus maintains 96.7\% of the original mAP.
We observe diminishing returns on the number of tokens and a greater decrease in mAP for \(T > 100 \; \text{ms}\), showing that $T$ should be chosen with care.
Refractory periods allow one to greatly reduce the number of tokens with only a small reduction in accuracy.
This shows that refractory periods are a powerful tool in making event-vision models faster.

\section{Limitations \& Future Work}
\label{sec:limitations}
The current formulation of Spiking Patches requires careful tuning of $\sigma$.
Future work will investigate adaptive thresholding to reduce the sensitivity of the threshold.
We have shown the effectiveness of Spiking Patches for GNNs, PCNs, and Transformers on gesture recognition and object detection.
This suggests strong generality, but evaluation of additional network types and tasks is required to strengthen the generality.
We plan in future work to explore Spiking Patches for additional types of networks and for more tasks (e.g. optical flow and object tracking).

\section{Conclusions}
\label{sec:conclusion}
We formulate tokenization as a method for representing events and preserving asynchrony and spatial sparsity at the representation level.
To this end, we have introduced a tokenizer for event cameras: Spiking Patches.
We have combined Spiking Patches with a GNN, PCN, and Transformer and evaluated its speed and accuracy on 3 datasets for gesture recognition and object detection, comparing it to frames, voxels, as well as events.
Spiking Patches was found to match or even surpass the accuracy of the other representations, while improving inference times up to 3.4$\times$ over voxels and up to 10.4$\times$ over frames.
We analyzed Spiking Patches and found that it is able to preserve both asynchrony and spatial sparsity.
The analysis also highlighted that Spiking Patches is efficient and applicable to real-time applications.
All together, Spiking Patches represents a novel promising direction, and we believe that this work will encourage more research into tokenization for event cameras.

\bibliographystyle{IEEEtran}
\bibliography{IEEEabrv,root}

\end{document}